\documentclass[11pt]{article}
\usepackage[dvips]{graphicx}
\usepackage{amssymb,amsmath,color}
\usepackage{url}

\usepackage[
    pdfstartview={FitH},
    bookmarks=true,
    bookmarksnumbered=true,
    bookmarksopen=true,
    bookmarksopenlevel=\maxdimen,
    pdfborder={0 0 0},
    colorlinks=true,
    linkcolor = blue,
    citecolor=blue,
    urlcolor=blue,
    filecolor=blue,
    pdfauthor={Simon Lacoste-Julien, Mark Schmidt, Francis Bach},
    pdftitle={A simpler approach to obtaining an O(1/t) convergence rate for the projected stochastic subgradient method},
    pdfdisplaydoctitle=true
]{hyperref}

\title{A simpler approach to obtaining an $O(1/t)$ convergence rate for the projected stochastic subgradient method}

\author{ \textbf{Simon Lacoste-Julien, Mark Schmidt, and Francis Bach}\\ 
INRIA - Sierra project-team\\
D\'epartement d'Informatique de l'Ecole Normale Sup\'erieure \\
Paris, France 
}

\newcommand{\BEAS}{\begin{eqnarray*}}
\newcommand{\EEAS}{\end{eqnarray*}}
\newcommand{\BEA}{\begin{eqnarray}}
\newcommand{\EEA}{\end{eqnarray}}
\newcommand{\BEQ}{\begin{equation}}
\newcommand{\EEQ}{\end{equation}}
\newcommand{\BIT}{\begin{itemize}}
\newcommand{\EIT}{\end{itemize}}
\newcommand{\BNUM}{\begin{enumerate}}
\newcommand{\ENUM}{\end{enumerate}}
\newcommand{\BA}{\begin{array}}
\newcommand{\EA}{\end{array}}

\newcommand{\BlackBox}{\rule{1.5ex}{1.5ex}}  %

\def \E { \mathbb{E} }

\begin{document}
 \maketitle
 
 \begin{abstract}
 In this note, we present a new averaging technique for the projected stochastic subgradient method. By using a weighted average with a weight of $t+1$ for each iterate $w_t$ at iteration $t$, we obtain the convergence rate of $O(1/t)$ with both an easy proof and an easy implementation. The new scheme is compared empirically to existing techniques, with similar performance behavior.
 \end{abstract}

 \section{Introduction}
 We consider a strongly convex function $f$ defined on a convex set $K$. We denote  by $\mu$ its strong convexity constant. Following~\cite{shalev2007pegasos,nemirovski2009robust,hazan2011beyond,rakhlin2012making}, we consider a stochastic approximation scenario where only unbiased estimates of subgradients of $f$ are available, with the projected stochastic subgradient method.
 
 More precisely, we assume that we have an increasing sequence of $\sigma$-fields $(\mathcal{F}_t)_{t \geqslant 0}$, such that 
 $w_0 \in K$ is $\mathcal{F}_0$-measurable and such that for all $t \geqslant 1$,
 \begin{equation} \label{eq:update}
 w_{t} = \Pi_K \big( w_{t-1} - \gamma_t g_t \big),
 \end{equation}
 where 
 
 \BIT
 \item[(a)]  $\Pi_K$ is the orthogonal projection on $K$, 
 \item[(b)] $ \E ( g_t | \mathcal{F}_{t-1} ) $ is almost surely a subgradient of $f$ at $w_{t-1}$ (which we denote $f'(w_{t-1})$), 
 \item[(c)]  $\E (\| g_t \|^2 ) \leqslant B^2 $ (finite variance condition).
 \EIT
 We denote by $w^\ast$ the unique minimizer of $f$ on $K$. 
 
 \section{Motivating example} \label{sec:example}
 Our main motivating example is the support vector machine (SVM) and its structured prediction extensions~\cite{shalev2010pegasos, tsochantaridis2006large,ratliff2007subgradient}, where the pairs $(x_t,y_t)$ for $t \geqslant 1$ are independent and identically distributed and $f(w) = \E \ell(y,w^\top x) + \frac{\mu}{2} \| w\|^2$, where $\ell(y,u)$ is a Lipschitz-continuous convex loss function (with respect to the second variable) and $K$ is the whole space (unconstrained setup). We then have 
$g_t = \ell'(y_t,w_{t-1}^\top x_t ) x_t + \mu w_{t-1}$, where $\ell'(y,u)$ denotes any subgradient with respect to the second variable.

If we make the additional assumption that $\E \| x\|^2$ is finite, then this setup satisfies the assumptions above with $B^2 = 4 L^2_{\ell} \E \| x\|^2$, where $L_{\ell}$ is the Lipschitz constant for $\ell$. We show this bound in Appendix~\ref{sec:SVMbound}.

Alternatively, we can consider $K$ to be a compact convex subset. This is used in particular in a projected version of the stochastic subgradient method for SVM in~\cite{shalev2007pegasos}. In this case, we can take $B^2 = (L_{\ell} \sqrt{\E \| x\|^2} + \mu \max_{w \in K} \| w\|)^2 $.
 
 \section{Convergence analysis}
 Following standard proof techniques~\cite{shalev2007pegasos,nemirovski2009robust}, we have:
 \BEAS
 \| w_t - w^\ast \|^2 & \leqslant & \| w_{t-1} - \gamma_t g_t - w^\ast \|^2 
 \mbox{ because orthogonal projections contract distances,}\\
  & = &  \| w_{t-1}- w^\ast \|^2 + \gamma_t^2 \| g_t \|^2 - 2 \gamma_t ( w_{t-1}- w^\ast)^\top g_t \\
 \E ( \| w_t - w^\ast \|^2  | \mathcal{F}_{t-1} ) 
 & \leqslant &  \| w_{t-1}- w^\ast \|^2 + \gamma_t^2 \E (\| g_t \|^2 | \mathcal{F}_{t-1} ) - 2 \gamma_t ( w_{t-1}- w^\ast)^\top f'(w_{t-1}) \\
 & \leqslant &  \| w_{t-1}- w^\ast \|^2 + \gamma_t^2 \E (\| g_t \|^2 | \mathcal{F}_{t-1} ) - 2 \gamma_t 
 \big[
 f(w_{t-1})  - f(w^\ast) + \frac{\mu}{2} \| w_{t-1}- w^\ast \|^2
 \big]
.
 \EEAS
 
 The last inequality is obtained from the $\mu$-strong convexity of $f$.
 Thus, by re-arranging the function values on the LHS and taking expectations on both sides, we get:
 \BEA
 2 \gamma_t \big[ \E f(w_{t-1}) - f(w^\ast)  \big]
 & \leqslant &  \gamma_t^2 \E \| g_t \|^2 + ( 1 - \mu \gamma_t ) \E \| w_{t-1}- w^\ast \|^2
 -\E \| w_t - w^\ast \|^2 \nonumber
\\
 \E f(w_{t-1}) - f(w^\ast)   
 & \leqslant &  \frac{\gamma_t B^2}{2} + \frac{ \gamma_t^{-1} - \mu  }{2} \E \| w_{t-1}- w^\ast \|^2
 - \frac{ \gamma_t^{-1}   }{2} \E \| w_t - w^\ast \|^2. \label{eq:ineq}
 \EEA
 
 \subsection{Classical analysis}
 
 With $\displaystyle \gamma_t = \frac{1}{\mu t}$, then inequality~\eqref{eq:ineq} becomes 
 $$
  \E f(w_{t-1}) - f(w^\ast)   
  \leqslant    \frac{  B^2}{2 \mu t} + \frac{\mu(t-1) }{2} \E \| w_{t-1}- w^\ast \|^2
 - \frac{ \mu t   }{2} \E \| w_t - w^\ast \|^2,
 $$
 and by summing from $t=1$ to $t=T$, we obtain:
 \BEAS
 \E f \bigg(
  \frac{1}{T} \sum_{t=1}^T w_{t-1}
 \bigg) - f(w^\ast) & \leqslant & 
 \frac{1}{T} \sum_{t=1}^T 
  \E f(w_{t-1}) - f(w^\ast)   \\
  &
  \leqslant &  \frac{B^2}{2 \mu T } \sum_{t=1}^T \frac{1}{t} + \frac{\mu}{2T} \left[0 -  
   T \E \| w_T- w^\ast \|^2 \right] \leqslant \frac{B^2}{2 \mu T } ( 1 + \log T).
 \EEAS
The first line used the convexity of $f$; the second line is obtained from a telescoping sum. We also obtain $\displaystyle \E \| w_T- w^\ast \|^2 \leqslant \frac{B^2}{\mu^2 T } ( 1 + \log T)$.

 \subsection{New analysis} \label{sec:new_analysis}

 With $\displaystyle \gamma_t = \frac{2}{\mu(t+1)}$ and multiplying inequality~\eqref{eq:ineq} by $t$, we obtain:
 
 \BEAS
 t \big[ \E f(w_{t-1}) - f(w^\ast)   \big]
 & \leqslant &  \frac{t B^2}{
\mu(t+1)} + 
\frac{\mu}{4} \bigg[ t (t-1)
  \E \| w_{t-1}- w^\ast \|^2
 - t(t+1) \E \| w_t - w^\ast \|^2
 \bigg]
\\
 & \leqslant &  \frac{  B^2}{
\mu } + 
\frac{\mu}{4} \bigg[ t (t-1)
  \E \| w_{t-1}- w^\ast \|^2
 - t(t+1) \E \| w_t - w^\ast \|^2
 \bigg].
 \EEAS
 
 By summing from $t = 1$ to $t=T$ these $t$-weighted inequalities, we obtain a similar telescoping sum, but this time the term with $B^2$ stays constant across the sum:
 
 \BEA
\sum_{t=1}^T  t \big[ \E f(w_{t-1}) - f(w^\ast)   \big]
   & \leqslant &  \frac{  T B^2}{
\mu } + 
\frac{\mu}{4} \bigg[ 0 
 - T(T+1) \E \| w_T - w^\ast \|^2
 \bigg]. \label{eq:telescope}
 \EEA
 
 Thus
 $$
 \E f \bigg( \frac{2}{T(T+1)} \sum_{t=0}^{T-1} (t+1) w_t \bigg) - f(w^\ast) 
 + \frac{\mu }{2} \E \| w_T - w^\ast \|^2
 \leqslant \frac{  2 B^2}{ \mu( T + 1 )}
 $$
 which implies 
 $$ \E f \bigg( \frac{2}{T(T+1)} \sum_{t=0}^{T-1} (t+1) w_t \bigg)   - f(w^\ast)  
 \leqslant
 \frac{  2 B^2}{ \mu( T + 1 )}$$
 and 
 $$
 \E \| w_T - w^\ast \|^2 \leqslant \frac{  4 B^2}{ \mu^2( T + 1 )}.
 $$
 
So by using the weighted average $\bar{w}_T \doteq \frac{2}{(T+1)(T+2)}\sum_{t=0}^T (t+1) w_t$ instead of a uniform average, we get a $O(\frac{1}{T})$ rate instead of $O(\frac{\log T}{T})$. Note that these averaging schemes are efficiently implemented in an online fashion as:
\begin{equation} \label{eq:wavg}
\bar{w}_t = (1-\rho_t) \bar{w}_{t-1}  + \rho_t w_t .
\end{equation}
For the proposed weighted averaging scheme, $\rho_t = 2/(t+2)$ (compare with $\rho_t = 1/(t+1)$ for the uniform averaging scheme).
 
 \section{Experiments}

\begin{figure}
\centering
\mbox{%
\includegraphics[width=.33\textwidth]{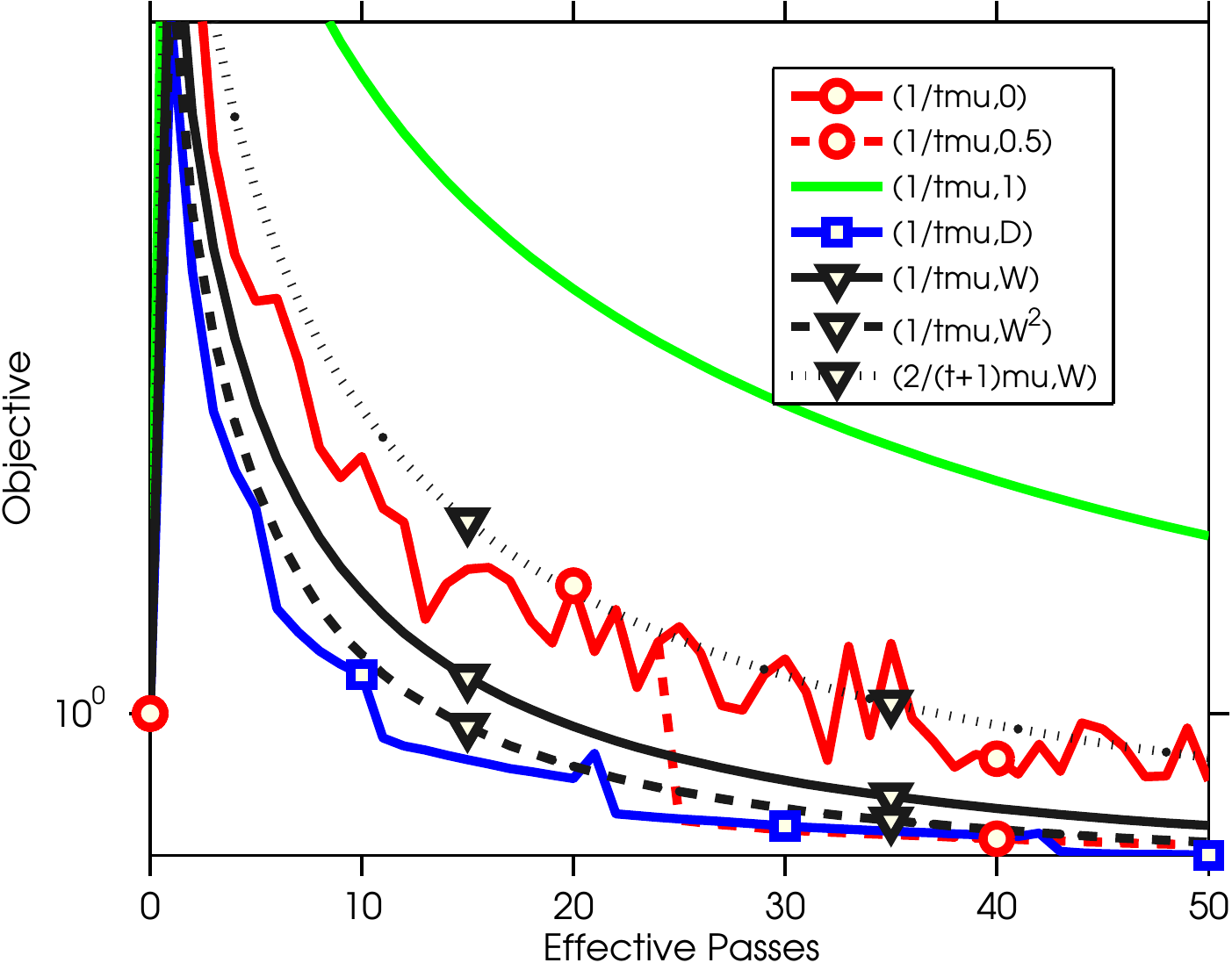} \hspace*{-.1cm}
\includegraphics[width=.33\textwidth]{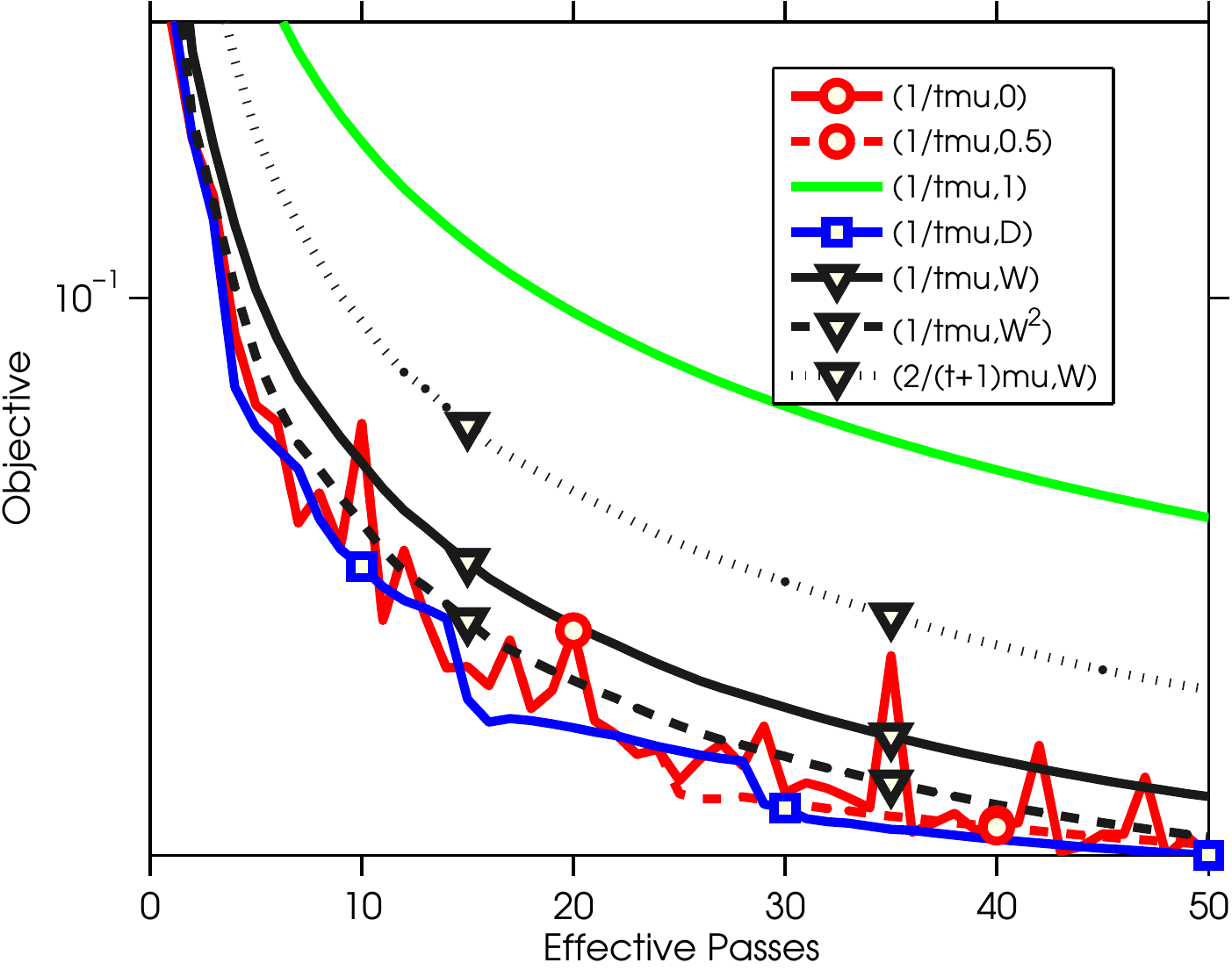} \hspace*{-.1cm}
\includegraphics[width=.33\textwidth]{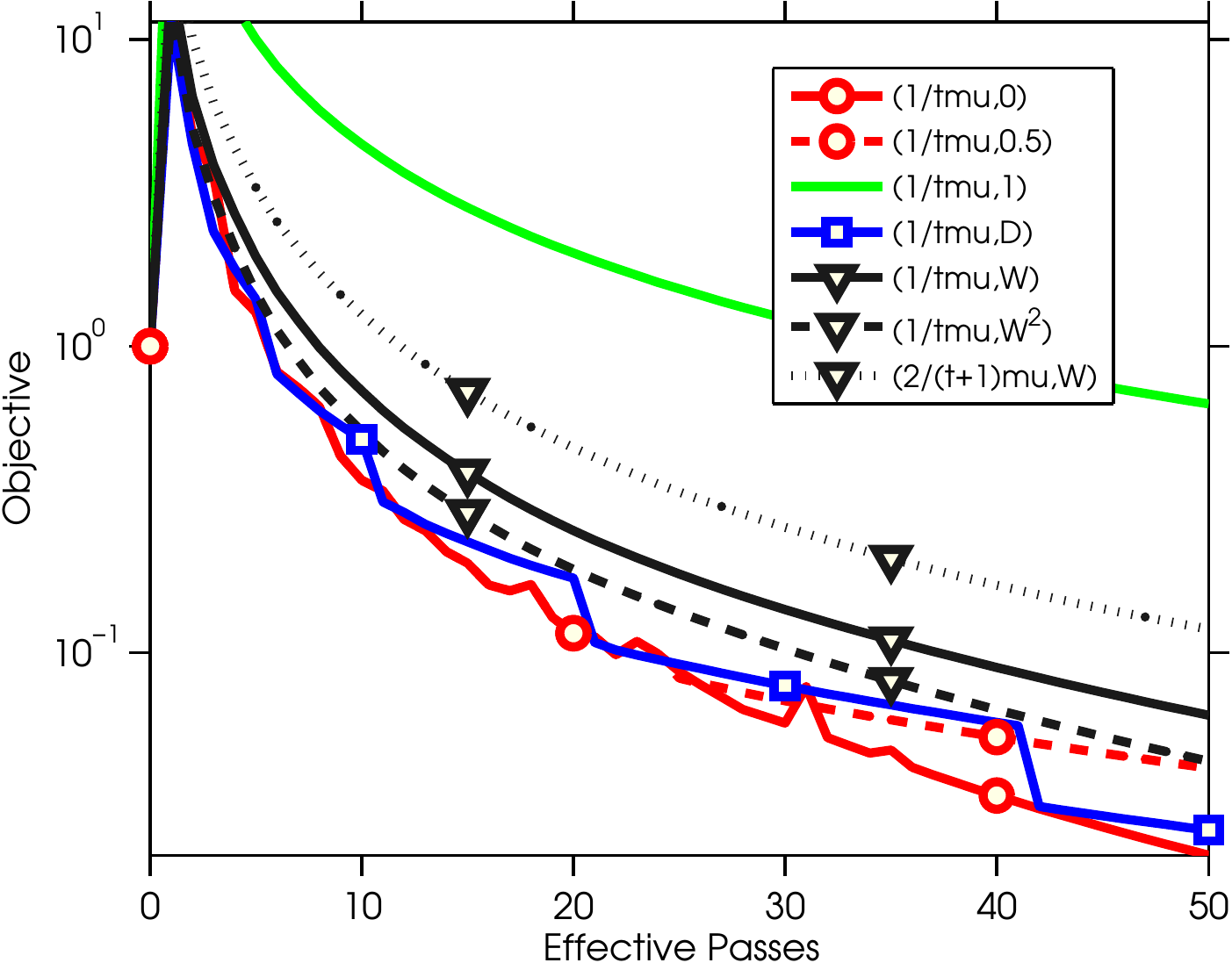} \hspace*{-.1cm}}
\mbox{%
\includegraphics[width=.33\textwidth]{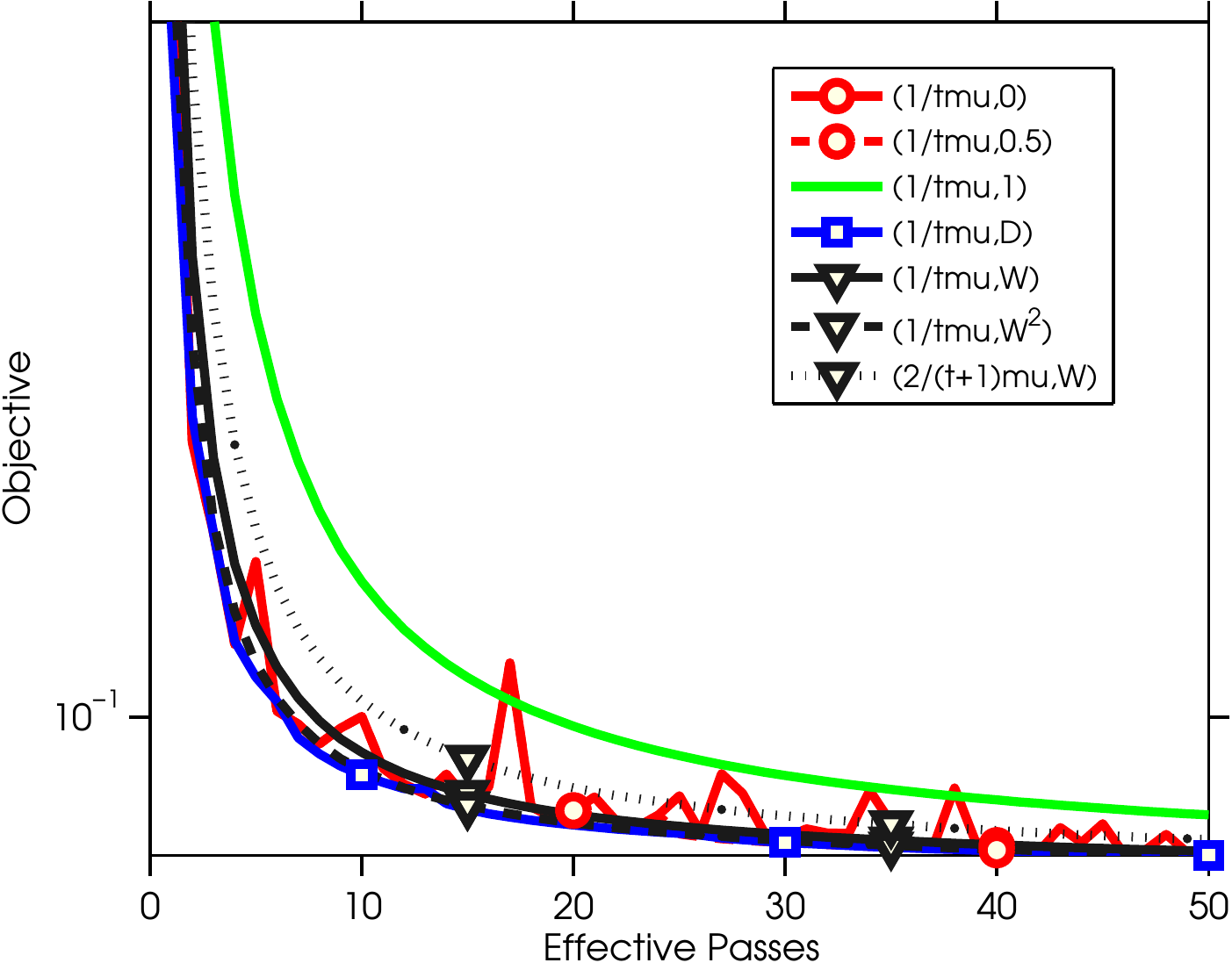} \hspace*{-.1cm}
\includegraphics[width=.33\textwidth]{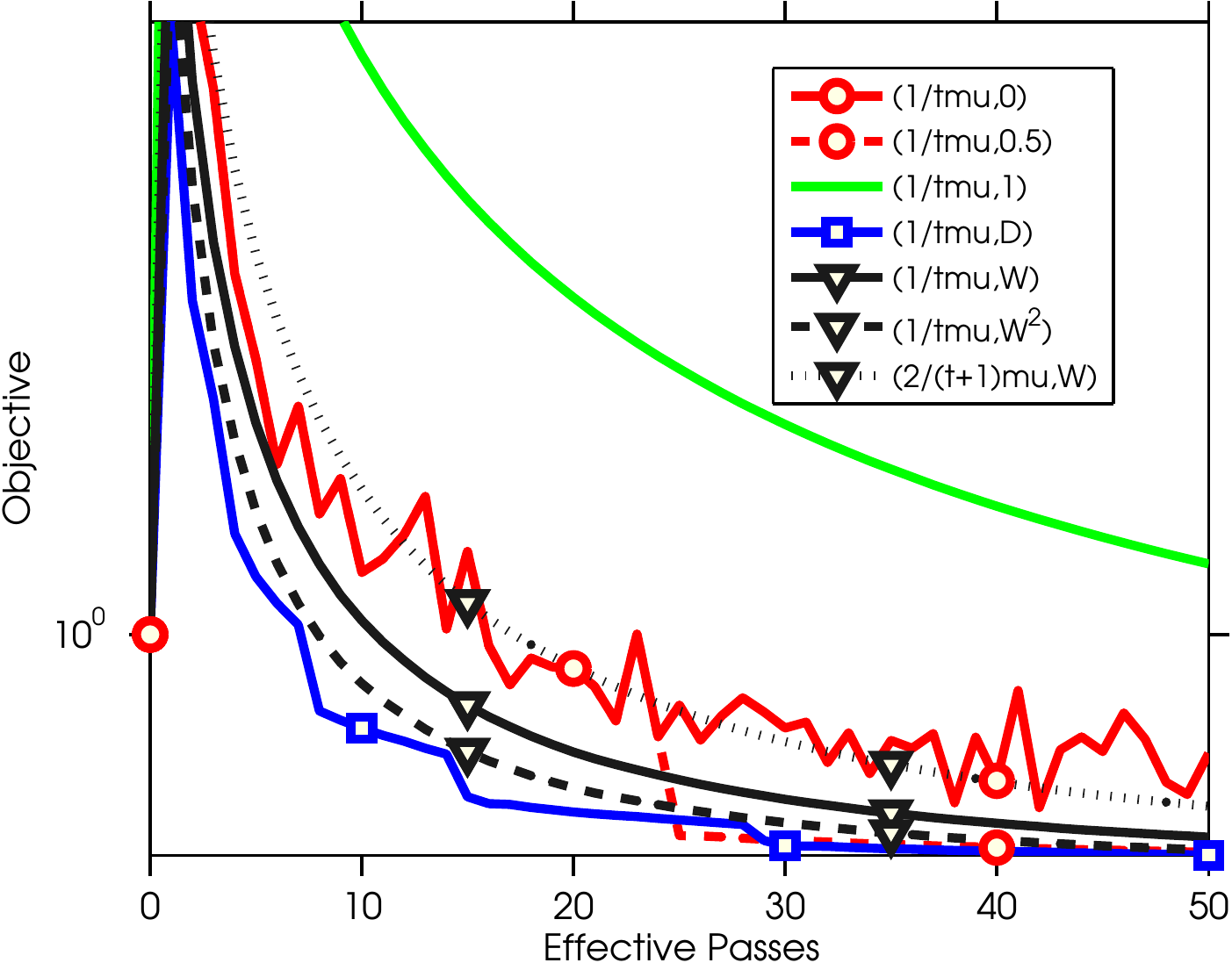} \hspace*{-.1cm}
\includegraphics[width=.33\textwidth]{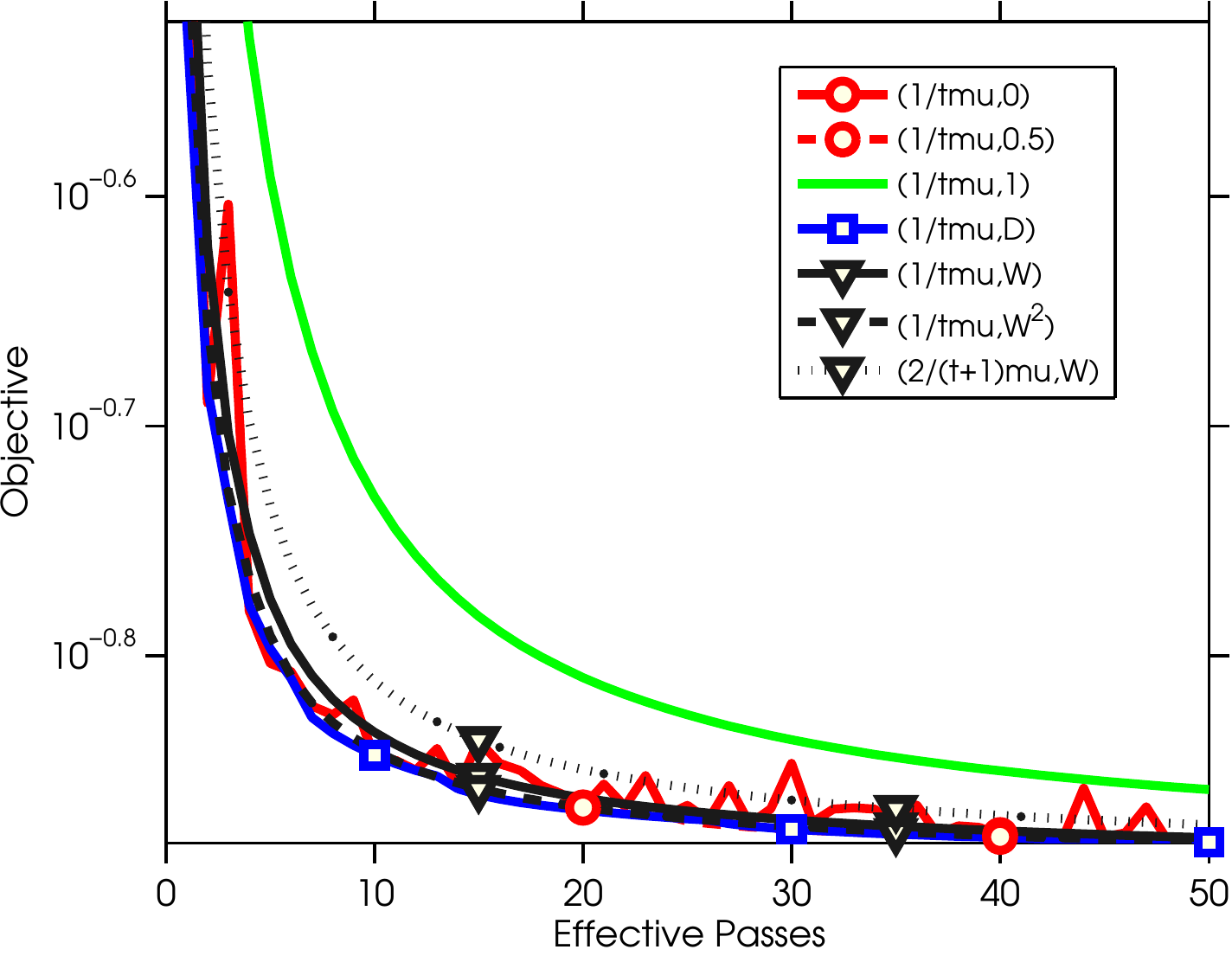} \hspace*{-.1cm}}
\caption{Comparison of optimization strategies for support vector machine objective. Top from to right: \emph{quantum}, \emph{protein}, and \emph{sido} data sets. Bottom from left to right: \emph{rcv1}, \emph{covertype}, and \emph{news} data sets. This figure is best viewed in colour.}
\label{fig:svm}
\end{figure} 
 
To test the empirical performance of the averaging scheme, we performed a series of experiments using the support vector machine optimization problem
\[
\min_w \; \frac{\lambda}{2}\|w\|^2 + \frac{1}{n}\sum_{i=1}^n \max\{0,1-y_iw^\top x_i\},
\]
where $x_i$ is in an Euclidean space and $y_i \in \{-1,1\}$.

We performed experiments on a set of freely available benchmark binary classification data sets.  The 
\emph{quantum} ($n=50 000$, $p=78$) and 
\emph{protein}   ($n=145 751$, $p=74$) data sets  were obtained from the KDD Cup 2004 website,\footnote{\small\url{http://osmot.cs.cornell.edu/kddcup}} the \emph{sido} data set ($n=12678$, $p=4932$) was obtained from the Causality Workbench website,\footnote{\small\url{http://www.causality.inf.ethz.ch/home.php}}
 while the \emph{rcv1}  ($n= 20 242$, $p=47 236$), \emph{covertype}   ($n=581 012 $, $p=54$), and \emph{news} ($n=19 996$, $p= 1 355 191$) data sets were obtained from the LIBSVM data website.\footnote{\small\url{http://www.csie.ntu.edu.tw/~cjlin/libsvmtools/datasets}} 
We added a (regularized) bias term to all data sets, and for dense features we standardized so that they would have a mean of zero and a variance of one. We set the regularization parameter $\lambda$ to $1/n$, although we found that the relative performance of the methods was not particularly sensitive to this choice. We didn't use any projection ($K$ is the whole space). Our experiments compared the following averaging strategies:
\BIT
\item[--] {\bf 0}:  No averaging.
\item[--] {\bf 1}: Averaging all iterates with uniform weight.
\item[--] {\bf 0.5}: Averaging the second half of the iterates with uniform weight, as proposed in~\cite{rakhlin2012making}.
\item[--] {\bf D}: Averaging all iterates since the last iteration that was a power of 2 with uniform weight (the `doubling trick'), also proposed in~\cite{rakhlin2012making}.
\item[--] {\bf W}: Averaging all iterates with a weight of $t+1$, as discussed in this note.
\item[--] {\bf W$^2$}: Averaging all iterates with a weight of $(t+1)^2$, which puts even further emphasis on recent iterations.
\EIT
We plot the performance of these different averaging strategies in Figure~\ref{fig:svm}, which shows the objective function against the number of effective passes through the data (the number of iterations divided by $n$). This figure uses a step size of $1/\mu t$ for all methods as we found this gave better performance than a step size of $2/\mu(t+1)$, although we include the performance of {\bf W} with the latter step-size for comparison.
In Figure~\ref{fig:svm}, we observe the following trends:
\BIT
\item[--] {\bf 0}: Not averaging at all is typically among the worst strategies. However, this proved to be the best strategy on the \emph{sido} data set. This may be because the method is still far from the solution after $50$ passes through the data.
\item[--] {\bf 1}: Uniform averaging of all iterates is always the worst strategy.
\item[--] {\bf 0.5}: Uniform averaging of the second half of the iterates is typically among the best strategies, provided we are in fact in the second half of the iterates.
\item[--] {\bf D}: The doubling trick typically gave among the best performance across the methods.
\item[--] {\bf W}: The proposed weighting typically performed between the doubling trick and not averaging.
\item[--] {\bf W$^2$}: Weighting the iterates by $(t+1)^2$ always outperformed weighting them by $t+1$.
\EIT

 \section{Discussion} 
 \BIT
 \item[--] We note that the averaging of linear approximations of $f$ (rather than the iterates) by $t+1$ is also used in the optimization strategy of Nesterov~\cite{nesterov2005smooth}, which achieves an optimal $O(1/t^2)$ convergence rate for optimizing (deterministic) objectives with Lipschitz-continuous gradients (see step 3 for their Equation 3.11). 
 \item[--] There are previous approaches to removing the $\log t$ term~\cite{hazan2011beyond,rakhlin2012making}, but the one presented in this note is arguably somewhat simpler to implement and analyze. Rakhlin et al. propose in~\cite{rakhlin2012making} the `1/2-suffix averaging' scheme (and the `doubling trick' that we used in the experiments). Their proof technique requires separately bounding $\E \| w_t-w^\ast\|^2$ and then controlling the sum of inequalities in~\eqref{eq:ineq} by using that only the last half of the iterations is averaged (the `1/2-suffix'). Hazan and Kale propose in~\cite{hazan2011beyond} the epoch-GD scheme, which uses a similar averaging schedule as in the `doubling trick' of~\cite{rakhlin2012making}, but using a fixed step-size within each geometrically sized `epoch' of averaging, as well as using the previous average as the initialization for an epoch.
 \item[--] We note that all the schemes presented in the experiments can have their convergence rate proven. Schemes \textbf{0} and \textbf{1} have $O((\log t)/ t)$ rate whereas the schemes \textbf{0.5}, \textbf{D}, \textbf{W} and \textbf{W$^2$} have $O(1/t)$ rate. We can show the $O(1/t)$ rate for general weighted averaging schemes (with weight $t^k$ for iterate $t$ for some fixed $k \ge 1$) as well as step-sizes of the form $\gamma_t = c/(t+b)$ for $c > 1/2$ and $b\geq 0$. The proof becomes longer though as the nice telescoping sum in~\eqref{eq:telescope} doesn't cancel out in these cases. One has to use instead a bound on $\E \| w_t-w^\ast\|^2$ such as in Lemma 1 in~\cite{rakhlin2012making} to control the non-canceling terms. The overall rate is still $O(1/t)$, but with different constants depending on $c$ and $k$. %
 \item[--] At the same time that we first posted this note, Shamir and Zhang independently proposed a similar weighted average scheme in~\cite{shamir2012sgd} which they call `polynomial-decay averaging'. They consider a running average scheme as in~\eqref{eq:wavg}, but with the more general $\rho_t = \frac{1+\eta}{t+1+\eta}$, where the integer $\eta \geq 0$ parameterizes the different schemes.\footnote{We note that the index $t$ is shifted by one between this note and their paper as their initial point is $w_1$ whereas ours is $w_0$. We also note that they use the misnomer `gradient descent' for their algorithm despite using subgradients which don't necessarily yield a descent direction.} $\eta = 0$ yields the standard uniform averaging scheme, whereas $\eta = 1$ yields the simple weighted average analyzed in Section~\ref{sec:new_analysis}. The general $\eta$ gives a weight of $O(t^\eta)$ for each iterate, similar to what was mentioned in the previous paragraph, but with a different exact formula. They provide in~\cite{shamir2012sgd} a proof of a rate of $O(1/t)$ for $\eta \geq 2$. The proof that we give in Section~\ref{sec:new_analysis} can be seen as complementary and is especially much simpler (as well as giving a tighter constant). We also note that the rate of $O((\log t)/ t)$ for the last iterate $w_t$ (scheme \textbf{0} above) is proven for the first time in~\cite{shamir2012sgd}.
 \item[--] While this paper focuses on the non-smooth case, it is still interesting to relate results to the smooth case (see, e.g.,~\cite{bach2011non} and references therein), where in the strongly convex case, averaging with longer step sizes---i.e., of the form $t^{-\alpha}$ with $\alpha \in (1/2,1)$---leads to better and more robust rates. Can larger step sizes improve results for the non-smooth case?
 \EIT
 
 \appendix
 
 \section{Finite variance bound for SVM} \label{sec:SVMbound}
 We derive here the finite variance bound $\E \| g_t \|^2  \leq  4 L^2_{\ell} \E \| x\|^2 = B^2$ for the general SVM-like objective considered in Section~\ref{sec:example} and update rule~\eqref{eq:update}. To see this, we consider the more general case of $f(w) = \E h(z,w) + \frac{\mu}{2} \| w\|^2$, where $h(z,w)$ is convex in $w$ for each $z$ (for SVM, $z=(x,y)$ and $h(z,w) = \ell(y,w^\top x)$). We make a Lipschitz-like (in expectation) assumption on $h$ that $\E \| h'(z,w) \|^2 \leq L^2$, where $h'(z,w)$ denotes any subgradient with respect to the second variable (note that $L^2 = L^2_{\ell} \E \| x\|^2$ for SVM). With $g_t = h'(z_t,w_{t-1})+\mu w_{t-1}$, Leibniz rule yields $\E (g_t | w_{t-1}) = f'(w_{t-1})$, as required by our setup (see~(1.3) in~\cite{nemirovski2009robust} for some regularity conditions for this to be true). Given this definition of $g_t$, we use the Minkowski inequality on the norm function\footnote{Minkowski inequality says that $\sqrt{\E (X+Y)^2} \leq \sqrt{\E X^2} + \sqrt{\E Y^2}$ for scalar random variables $X$ and $Y$. If we have $a=b+c$ for some random vectors $a,b,c$, by the triangle inequality, we have $\|a\|\leq\|b\|+\|c\|$ and so $\sqrt{\E \|a\|^2} \leq \sqrt{\E (\|b\|+\|c\|)^2} \leq \sqrt{\E \|b\|^2} + \sqrt{\E \|c\|^2}$ by using Minkowski on the norm of the random vectors.} to get:
 $$
 	\sqrt{\E \| g_t \|^2} \leq \sqrt{\E \| h'(z_t,w_{t-1}) \|^2} + \mu \sqrt{\E \| w_{t-1} \|^2}
 	\leq L + \mu \sqrt{\E \| w_{t-1} \|^2}.
 $$
We can then obtain the required bound of $(2L)^2$ on $\E \| g_t \|^2$ by showing that $\sqrt{\E \| w_{t-1} \|^2} \leq L/\mu$. This can easily be proven by induction, with the assumption that $\gamma_t \leq 1/\mu$ and either $\gamma_1 = 1/\mu$ or $\sqrt{\E \|w_0 \|^2} \leq L/\mu$ (these assumptions are satisfied by the step sizes considered in this note). To see this, we use the subgradient update~\eqref{eq:update} applied to this form of $f(w)$:
 \BEAS
w_t & = & (1-\mu \gamma_t) w_{t-1} - \gamma_t h'(z_t, w_{t-1}). \\
 \EEAS
Applying Minkowski inequality again, we get
 \BEAS
\sqrt{\E \| w_t \|^2} & \leq &(1-\mu \gamma_t) \sqrt{\E \| w_{t-1} \|^2}+ \gamma_t \sqrt{\E \| h'(z_t,w_{t-1} \|^2} \\
& \leq & (1-\mu \gamma_t) \sqrt{\E \| w_{t-1} \|^2} + \mu \gamma_t \frac{L}{\mu}.
 \EEAS
The first line above used the assumption that $\gamma_t \leq 1/\mu$ to ensure that $(1-\mu \gamma_t)$ is non-negative. The assumption $\gamma_1 = 1/\mu$ or $\sqrt{\E \|w_0 \|^2} \leq L/\mu$ then yields the base case of $t=1$. Plugging in the induction hypothesis then yields:
 \BEAS
\sqrt{\E \| w_t \|^2} & \leq &(1-\mu \gamma_t) \frac{L}{\mu} + \mu \gamma_t \frac{L}{\mu} = \frac{L}{\mu},
 \EEAS
which completes the proof.
  \bibliography{no_pain}
\bibliographystyle{unsrt}

   \end{document}